\documentclass[runningheads]{llncs}
\usepackage{graphicx}
\usepackage{multirow}
\usepackage{color, colortbl}
\begin{document}
\title{On the Temporality of Priors in Entity Linking}
\author{Renato Stoffalette Jo\~{a}o\inst{1}\orcidID{0000-0003-4929-4524}}
\authorrunning{R. Stoffalette Jo\~{a}o}
\institute{L3S Research Center\\Appelstra{\ss}e 9A - 30167 \\ Hannover, Germany\\
\email{joao@L3S.de}}
\maketitle              % typeset the header of the contribution
\begin{abstract}
Entity linking is a fundamental task in natural language processing which deals with the lexical ambiguity in texts. An important component in entity linking approaches is the mention-to-entity \emph{prior} probability. Even though there is a large number of works in entity linking, the existing approaches do not explicitly consider the time aspect, specifically the {\em temporality} of an entity's prior probability. We posit that this prior probability is temporal in nature and affects the performance of entity linking systems. In this paper we systematically study the effect of the prior on the entity linking performance over the temporal validity of both texts and KBs. 

%Does the entity popularity  change  over  time  ?   If  so,  how  often does it occur and how much does it affect the overall named entity disambiguation task ?  Given weonly take into consideration the top 5 ranked can-didate entities for each entity mention, how often does the top candidate entity probability change ?

%In this paper, we validate the prior probability of candidate entities is temporal in nature 

\end{abstract}
\section{Introduction and Motivation}
\label{lab:intro}
Entity linking is a well studied problem in natural language processing which involves the process of identifying ambiguous entity mentions (i.e persons, locations and organisations) in texts and linking them to their corresponding unique entries in a reference knowledge base. There has been numerous approaches and eventually systems proposing solutions to the task at hand. To mention a few, AIDA~\cite{Hoffart2011}, Babelfy~\cite{journals/tacl/0001RN14}, WAT~\cite{Piccinno2014} and AGDISTS~\cite{Usbeck2014} for example, rely on graph based algorithms and the most recent approaches rely on techniques such as deep neural networks~\cite{HeLLZZW13} and semantic embeddings~\cite{HuangHJ15,Zwicklbauer2016}. 

An important component in most approaches is the probability that a mention links to one entity in the knowledge base. The prior probability, as suggested by Fader et al.~\cite{Fader09}, is a strong indicator to select the correct entity for a given mention, and consequently adopted as a baseline. Computation of this prior is typically done over knowledge sources such as Wikipedia. Wikipedia in fact provides useful features and has grounded several works on entity linking~\cite{bunescu-etal-2006,Cucerzan,Mihalcea2007,Milne2008b,Hoffart2011}.%\tocheckr{ Wikipedia as an organization has not created several named entity linkers. Authors of other organizations have used Wikipedia features in their named entity linkers.}

An entity's popularity is temporally sensitive and may change due to short term events. Fang and Chang~\cite{FangC14} noticed the probability of entities mentioned in texts often change across time and location in micro blogs, and in their work they modeled spatio-temporal signals for solving ambiguity of entities. We, on the other hand, take a macroscopic account of time, where perceivably a larger fraction of mention to entity bindings might not be observable in the short time duration but are only evident over a longer period of time, i.e., over a year. These changes might be then reflected in a reference knowledge base and disambiguation methods can produce different results for a given mention at different times.
% On the other hand, a larger fraction of mention to entity bindings might not be observable in the short time duration but are only evident when we take a macroscopic account of time, i.e., over a year. Consequently, we take 

When using a 2006 Wikipedia edition as a reference knowledge base for example, the mention  \textit{Amazon} shows different candidates as linking destinations, but the most popular one is the entity page referring to \textit{Amazon River}, whilst when using a 2016 Wikipedia edition, the same term leads to the page about the e-commerce company \textit{Amazon.com} as the most popular entity to link to.

In this paper, we systematically study the effect of temporal priors on the disambiguation performance by considering priors computed over snapshots of Wikipedia at different points in time. We also consider benchmarks that contain documents created and annotated at different points in time to better understand the potential change in performance with respect to the temporal priors.

We firstly show that the priors change over time and the overall disambiguation performance using temporal priors show high variability. This strongly indicates that temporal effects should be not only taken into account in (a) building entity linking approaches, but have major implications in (b) evaluation design, when baselines that are trained on temporally distant knowledge sources are compared.

\section{Problem Definition}

In  this  section we briefly define the entity linking task as well as describe our methodology.
%\section{Terminology and %Problem Definition}
%
%Table~\ref{tab:terminology} shows the terminology used throughout the paper.
%\begin{table}[!h]
%\resizebox{\columnwidth}{!}{%
%\begin{tabular}{l p{7cm}l}	
%\hline 
%	\textbf{\textit{t}} & A fixed time period.\\
%	\textbf{\textit{W$^{t}$}} & A knowledge base at time \textit{t}.\\
%	\textbf{\textit{D$^{t}$}}  &  A set of documents at time \textit{t}.\\%, where (\textbf{\textit{D}} = \{\textbf{$d_{t1}$}, \textbf{$d_{t2}$}, $\ldots$,	 \textbf{$d_{tn}$}\}).\\
%	\textit{\textbf{d$^{t}$}} &  A document at time \textit{t} and \textit{d$^{t}$} $\in$ \textit{D$^{t}$}.\\
%	\textbf{\textit{e$^{t}$}} & An entity referring to a Wikipedia article.\\
%	\textit{\textbf{E$^{t}$}} & A set of entities, \textit{E$^{t}$} = \{$e^{t}_{1}$, $e^{t}_{2}$, \ldots, $e^{t}_{n}$ \} extracted from \textit{W$^{t}$}.\\
%	\textbf{\textit{m$^{t}$}} & A mention represented by a single term or a sequence of terms extracted from \textit{d$^{t}$}.\\
%	\textit{\textbf{M$^{t}$}} & A set of mentions, \textit{M$^{t}$} = \{{$m^{t}_{1}$}, $m^{t}_{2}$, \dots, $m^{t}_{n}$ \}.\\ 
%	\textit{\textbf{P(e$^{t}$)}}  &  The probability of $e^{t}$.\\
%	\textit{\textbf{P(e$^{t}|m^{t}$)}} 	  &  The probability of $e^{t}$ given $m^{t}$.\\
%	\hline
%\end{tabular}
%}
%\caption {Terminology.} 
%\label{tab:terminology} 
%\end{table}
%We formalize the named entity disambiguation problem as follows. 

Consider a document \textit{d} from a set of documents \textit{D} = \{$d_{1}$, $d_{2}$, $\ldots$, $d_{n}$\}, and a set of mentions \textit{M} = \{$m_{1}$, $m_{2}$, $\ldots$, $m_{n}$ \} extracted from \textit{d}. The goal of the entity linking is to find a unique identity represented by an entity \textit{e} from a set of entities \textit{E} = \{$e_{1}$, $e_{2}$, $\ldots$, $e_{n}$\}, with relation to each mention \textit{m}. The set of entities \textit{E} is usually extracted from a reference knowledge base \textit{KB}.    

A typical entity linking system generally performs the following steps: 1) mention detection which extracts terms or phrases that may refer to real world entities, and 2) entity disambiguation which selects the corresponding knowledge-base KB entries for each ambiguous mention.

Since we take into account the time effect on the disambiguation task, we now pose entity linking at a specific time \textit{t} as follows. Given a document \textit{d$^{t}$} $\in $\textit{D$^{t}$} and a set of mentions \textit{M} = \{{$m_{1}$}, $m_{2}$, \ldots, $m_{n}$ \} from document \textit{d$^{t}$}, the goal of the entity linking at time \textit{t} is to find the correct mapping entity \textit{e$^{t}$} $\in$ \textit{E$^{t}$} with relation to the mention \textit{m}. The difference now is that the set of entities \textit{E$^{t}$}
is extracted from the reference knowledge base \textit{KB} at different time periods. %\\[-1.5cm]

\subsection{Candidate Entities Generation and Ranking}
As suggested by Fader et al.~\cite{Fader09}, the entity's prior probability is a strong indicator to select the correct entity for a given mention. In our case the entity's prior probability is directly obtained from the Wikipedia corpus. 
%We used  the pages-articles.xml  thatcontains current version of all article pages, templates, andother pages. The parser takes as input the Wikipedia dumpﬁle and analyzes the content enclosed in the various XMLtags. We use the Wikipedia dump in July 2011. In the ﬁrstrunning parse, it builds the set of redirection pairs (i.e. onearticle is a pointer to some actual one), list of pages, dis-ambiguation pages, and set of titles of all articles in thatWikipedia edition. 
To calculate entities' probability, we parsed all the articles from a Wikipedia corpus and collected all terms that were inside double square brackets in the Wikipedia articles.  [[\textit{Andy Kirk (footballer)} $|$ \textit{Kirk}]] for instance, represents a pair of mention and entity where \textit{Kirk} is the mention term displayed in the Wikipedia article and \textit{Andy Kirk (footballer)} is the title of the Wikipedia article corresponding to the real world entity.
%In this way we created a reference knowledge base of mentions and their referring entities.
In this way we created a list of mentions and possible candidate entities according to each Wikipedia snapshot used in this paper.

The probability of a certain entity \textit{e$^{t}$} given a mention \textit{m} was only calculated if the entity had a corresponding article inside Wikipedia at time \textit{t}. Thus, the probability \textit{P(e$^{t}|m$)} that a mention \textit{m} links to a certain entity \textit{e$^{t}$} is given by the number of times the mention \textit{m} links to the entity \textit{e$^{t}$} over the number of times that \textit{m} occurs in the whole corpus at time \textit{t}. 

We created dictionaries of mentions and their referring entities ranked by popularity of occurrence for every Wikipedia edition as seen on Table~\ref{tab:tabwiki}. As an example of mention and its ranked candidate entities, in the KB created from the 2016 Wikipedia edition, the mention \textit{Obama} refers in 86.15\% of the cases to the president \textit{Barack Obama}, 6.47\%  to the city \textit{Obama, Fukui} in Japan, 1.79\% to the genus of planarian species \textit{Obama (genus)}, and so on and so forth.

We filtered out mentions that occurred less than 100 times for simplicity matters in the whole corpus and for every mention we checked whether the referring candidate entities pointed to existing pages inside the Wikipedia corpus at a given time, and only after these steps we calculated the prior probability values the entities.

Our framework supports multiple selection of mention-entity dictionaries created from different \textit{KB}s based on Wikipedia snapshots from different years.

\begin{table}[t]
\small
\centering
\caption {Information about the Wikipedia editions used for mining mention and entities. \#Pages refers only to the number of entities' pages, excluding special pages.} 
\label{tab:tabwiki}
\begin{tabular}{|c|c|c|}	
\hline
\textbf{Year} & \textbf{Date} & \textbf{\#Pages} \\ \hline
2006   & 30/11/2006 & $\sim$ 1.4 M	\\ \hline
2008   & 03/01/2008 & $\sim$ 1.9 M	\\ \hline
2010   & 15/03/2010	& $\sim$ 2.8 M	\\ \hline
2012   & 02/09/2012 & $\sim$ 3.5 M	\\ \hline
2014   & 06/11/2014 & $\sim$ 4.1 M	\\ \hline
2016   & 01/07/2016 & $\sim$ 5.1 M	\\ \hline
\end{tabular}
\vspace{-3mm}
\end{table}

\section{Experiments and Results}

\subsection{Datasets}
In order to evaluate our experiments we employed some data sets that are widely used benchmark datasets for entity linking tasks.
\textit{ACE04} is a news corpus introduced by Ratinov et al.~\cite{Ratinov2011}
and it is a subset from the original ACE co-reference data set~\cite{doddington2004automatic}. \textit{AIDA/CONLL} is proposed by Hoffart et al.~\cite{Hoffart2011} and it is based on the data set from the CONLL 2003 shared task~\cite{sang2003introduction}. \textit{AQUAINT50} was created in the work proposed by Milne \& Witten~\cite{Milne2008b}, and is a subset from the original AQUAINT newswire corpus~\cite{graffaquaint}.
\textit{IITB} is a dataset extracted from popular web pages about sports, entertainment, science and technology, and  health\footnote{\url{http://news.google.com/}}\footnote{\url{http://www.espnstar.com/)}}, and it was created in the work proposed by Kulkarni et al.~\cite{Kulkarni2009}.
\textit{MSNBC} was introduced by Cucerzan~\cite{Cucerzan} and contains news documents from 10 MSNBC news categories.
Table~\ref{tab:datasets} shows more details about these datasets including the number of documents, documents' publication time, number of annotations as well as the reference knowledge base time.

\begin{table}[b]
\small
\centering
\caption{\textit{\#Docs} is the number of documents. \textit{Docs Year} is the documents' publication time. \textit{\#Annotations} is the number of annotations (Number of \textit{non-NIL} annotations). \textit{Annot. Year} is the reference KB time period where the annotations were taken from.}
\label{tab:datasets} 
\begin{tabular}{|l|c|c|c|c|}	
   \hline
   \textbf{Dataset} & \textbf{\#Docs} &  \textbf{Docs Year} &  \textbf{\#Annotations} & \textbf{Annot. Year} \\ \hline
		ACE04	~\cite{Ratinov2011} 						 		 & 57  	 & 2000 	   & 257    		 & 2010  \\ \hline
		AIDA/CONLL	~\cite{Hoffart2011}						 & 231 	 & 1996        & 4.485  		 & 2010  \\ \hline
		AQUAINT50	~\cite{Milne2008b}  							 & 50    & 1998-2000   & 727   	 		 & 2007  \\ \hline
		IITB	~\cite{Kulkarni2009}								 & 107   & 2008 	   & 12.099  	 & 2008  \\ \hline
		MSNBC	~\cite{Cucerzan}       						 		 & 20    & 2007        & 747		     & 2006  \\ \hline		
\end{tabular}
\vspace{-5mm}
\end{table}

\subsection{Prior Probability Changes}

In many entity linking systems, the entity mentions that should be linked are given as the input, hence the number of mentions generated by the systems equals the number of entity mentions that should be linked. For this reason most researchers use accuracy to evaluate their method's performance. Accuracy is a straightforward  measure calculated as the number of correctly linked mentions divided by the total number of mentions. 

Since we take into account the time variation, we only calculated accuracy over the total number of annotations that persisted across time, i.e. the entities from the ground truth that were also present in every Wikipedia edition used in this paper. Table~\ref{tab:titlebaseline1} shows the accuracy calculated on the ground truth datasets using the prior probability model
from different time periods.
We can observe an accuracy change from 77.19\% to 82.63\% on \textit{ACE04} using models created from Wikipedia 2006 and 2010 editions respectively, from
64.80\% to 69.07\%
for \textit{AQUAINT50} using models from 2006 and 2012 editions, from 64.13\% to 68.16\%
for \textit{AIDA/CONLL} using models from 2008 and 2014, from 46.60\% to 49.76\% on \textit{IITB} using models from 2014 and 2006, and for \textit{MSNBC} a change from  63.82\% to  65.86\% using models from Wikipedia 2012 and 2008 editions respectively

Even though it is out of the scope of this work to spot a temporal trend on the entities changes when using knowledge bases from different time periods, we can clearly see there is some temporal variability which is easily observed by the influence on the accuracy calculated over the ground truth datasets.  We observe that a simplistic popularity only based method that takes into account reference \textit{KB}s from different time periods can produce an improvement of 5.4 percentage points in the best case for the \textit{ACE04} dataset and 2.0 percentage points in the worst case for \textit{MSNBC}  dataset. 

%\tocheckr{There is no clear observable pattern of the model's accuracy with the different datasets and Wikipedia editions. What is it that the patterns show beyond some slight changes?}

\begin{table}[b]
\vspace{-2mm}
\label{tab:titlebaseline1}
\caption {Accuracy of the models on different data sets across different time periods.
}
\centering
\small
\vspace{-2mm}
  \begin{tabular}{|l|c|c|c|c|c|c|}	    \hline
	\textbf{Dataset} & \textbf{2006}   & \textbf{2008} & \textbf{2010} & \textbf{2012} & \textbf{2014} & \textbf{2016}   \\ \hline
		ACE2004      	    	& 77.19 & 81.17 & \textbf{82.63} & 80.96 & 80.54 & 79.49 \\ \hline
		AIDA/CONLL\_testb  	& 61.86 & 64.13 & 66.47 & 67.78 & \textbf{68.16} & 68.14 \\ \hline
		AQUAINT50  	 		& 64.80 & 68.18 & 68.92 & \textbf{69.07} & 67.30 & 66.86 \\ \hline
		IITB     				& \textbf{49.76} & 49.43 & 49.50 & 47.78 & 46.60 & 47.60 \\ \hline
		MSNBC    				& 65.30 & \textbf{65.86} & 65.67 & 63.82 & 64.56 & 65.67 \\ \hline
\end{tabular}
\vspace{-5mm}
\end{table}

\subsection{Comparing Ranked Entities}

We detected distinct changes when it comes to entity linking using Wikipedia as a knowledge base. The first case occurs when the entity page title changes but still refers to the same entity in the real world. For example in the 2006 Wikipedia edition the mention \textit{Hillary Clinton} showed higher probability of linking to the referring entity page titled \textit{Hillary Rodham Clinton} and in the 2016 Wikipedia edition, the same mention was most likely to be linked to the entity page titled \textit{Hillary Clinton}. In this case only the entity page title changed but they both refer to the same entity in the real world.

The second case happens when an entity's popularity actually changes over time. For example in the 2006 Wikipedia edition, the mention \textit{Kirk} was most likely to be linked to the entity page titled \textit{James T. Kirk} whereas in the 2016 Wikipedia edition the same mention showed a higher probability of linking to the entity page titled \textit{Andy Kirk (footballer)}.  

Another observation is the case when an entity mention that was considered unambiguous in the past and became ambiguous in a newer Wikipedia edition  due  to the  addition  of  new  information to Wikipedia. For example in the 2006 Wikipedia edition the mention \textit{Al Capone} showed a single candidate entity, the  
north american gangster and businessman \textit{Al Capone}, while in the newer 2016 Wikipedia edition, the same mention showed more candidate entities, including the former one plus a movie, a song, and other figures with the same name.
%For simplicity matters we filtered out mentions that occurred less than 100 times in the whole corpus and for every mention we checked whether the referring candidate entities pointed to existing pages inside the Wikipedia corpus. Only after these steps we calculated the prior probability values for an entity given a certain mention.

\subsubsection{Top Ranked Entity Changes}
%\tocheckr{How did you calculate the rank correlation? Usually a value of 1 means total agreement while a value of 0 means total disagreement, and a value of -1 means an in inverse behavior. Did you use a distance instead? This needs to be clarified. As it stands, I am not sure how to read the results.}
Initially we were only concerned with the top ranked candidate entity for each mention. We made comparisons between the dictionaries of mentions from Wikipedia editions 2006 and 2016 and despite the fact of observing 33,531 mentions in the 2006 version and 161,264 mentions in the 2016 version, only 31,123 mentions appeared in both editions. Moreover, when we take into consideration both the ambiguous and unambiguous mentions, in 9.44\% of the cases the mentions change their top ranked candidate entities, whilst when removing the unambiguous mentions this number increases to 15.36\%. This is mainly due to the fact that most of the unambiguous mentions keep the same entity bindings, even though we spotted cases of mentions that were unambiguous and became ambiguous in a more recent knowledge base.

%\begin{table}[h]
%\small
%\centering
%  \begin{tabular}{|l|c|}	    \hline
%  		\textit{\textbf{\#}} mentions in Wikipedia	2006   & 33.531  \\ \hline
%		\textit{\textbf{\#}} mentions in Wikipedia	2016   & 161.264 \\ \hline 
%		\textit{\textbf{\#}} mentions in common  		   & 31.123  \\ \hline
%		\textit{\textbf{\#}} mentions where top entity changed & 2.939 \textit{(9.44\%)} \\ \hline
%  \end{tabular}
%  \caption {Top ranked entity disambiguation results (All mentions).} 
% \label{tab:top1statistics}
%\end{table}

%\begin{comment}
%\begin{table}[h]
%\small
%\centering
%  \begin{tabular}{|l|c|}	    \hline
%  		\textit{\textbf{\#} }mentions in Wikipedia	2006   & 20.511 \\ \hline
%		\textit{\textbf{\#}} mentions in Wikipedia	2016   & 89.815 \\ \hline
%		\textit{\textbf{\#}} mentions in common  		   & 18.727  \\ \hline		
%		\textit{\textbf{\#}} mentions where top entity changed & 2.878 \textit{(15.36\%)} \\ \hline
%  \end{tabular}
%\caption {Top ranked entity disambiguation results (Only ambiguous mentions).} 
% \label{tab:top1statisticsII} 
%\end{table}
%\end{comment}

\subsubsection{Top 5 Entities Changes}
In another experiment we wanted to calculate the entities rank correlation. One way to calculate rank correlation for lists that do not have all the element in common, is to ignore the non conjoint elements, but unfortunately this approach is not satisfactory since it throws away information. Hence, a more satisfactory approach, as proposed by Fagin et al.~\cite{Fagin2003}, is to treat an element \textit{i} which appears ranked in list \textit{$L_{1}$} and does not appear in list \textit{$L_{2}$}, at position \textit{k+1} or beyond, considering \textit{$L_{2}$}'s depth is \textit{k}. This measure was used to assess the changes in the top 5 candidate entities rank positions.

We calculated the rank correlation for 18,727 mentions, since this is the number of mentions that are ambiguous and appears both in the 2006 and 2016 Wikipedia corpus. %Table~\ref{tab:kendall_20062016} shows this measure averaged for all mentions. 
We normalized our results so the values would lie between [0,1]. Any value close to 0 means total agreement while any value close to 1 means total disagreement. Thus we observed an average value of  0.59  with a variance of 0.05 and a standard deviation of  0.21. We noticed that in 71.98\% of the cases the rank correlation values are greater than 0.5. That tells us there is some significant number of changes in the candidate entities' rank's positions.
Table~\ref{tab:tabtop} shows the mention \textit{Watson} and its top 5 candidate entities together with their respective prior probabilities extracted from two different Wikipedia editions, one from 2006 and one from 2016.
%\tocheckr{Table 4 shows an interesting result, but you should comment on the results, if only with a single sentence.}

\begin{table}[!h]
\vspace{-3mm}
\small
\centering
\caption{A mention example and its top 5 ranked candidate entities captured from two Wikipedia editions.} 
\vspace{-5mm}
\begin{tabular}{llcl} \\ \hline
\textbf{Mention} & \textbf{Entity} & \textbf{\textit{P(e$^{t}$)}} & \textbf{Year} \\ \hline
			\multirow{5}{*}{Watson}
								 & \multicolumn{1}{l}{Doctor Watson } & \multicolumn{1}{l}{0.146} &  \\\cline{2-3}
								 & \multicolumn{1}{l}{James D. Watson } & \multicolumn{1}{l}{0.130 } &  \\\cline{2-3}
								 & \multicolumn{1}{l}{Watson, Australian Capital Territory} & \multicolumn{1}{l}{0.115 } & 2006 \\\cline {2-3}
								 & \multicolumn{1}{l}{Division of Watson} & \multicolumn{1}{l}{0.076} & \\\cline{2-3}
								 & \multicolumn{1}{l}{Watson} & \multicolumn{1}{l}{0.061} &   \\ \hline \hline
			\multirow{5}{*}{Watson}
								 & \multicolumn{1}{l}{Watson (computer)} & \multicolumn{1}{l}{0.068} &  \\\cline{2-3}
								 & \multicolumn{1}{l}{Ben Watson (footballer, born July 1985)} & \multicolumn{1}{l}{0.054 } &  \\\cline{2-3}
								 & \multicolumn{1}{l}{Je-Vaughn Watson} & \multicolumn{1}{l}{0.050} & 2016 \\\cline{2-3}
								 & \multicolumn{1}{l}{Jamie Watson (soccer)} & \multicolumn{1}{l}{0.047} &  \\\cline{2-3}
								 & \multicolumn{1}{l}{Arthur Watson (footballer, born 1870)} & \multicolumn{1}{l}{0.043} &  \\\cline{2-3} \hline
 \end{tabular}
\label{tab:tabtop}
\end{table}
\vspace{-5mm}

\vspace{-4mm}
\section{Conclusions and Future Work}
In this work we conducted experiments with different Wikipedia editions and also created an entity linking model that uses the entity's prior probability calculated over different Wikipedia snapshots. One limitation of previous works is the fact that the systems are trained on a fixed time Wikipedia edition.
An entity's prior probability is temporal in nature, and we have observed in our experiments that mention to entity bindings change over time. We could clearly see some temporal variability which should be taken into account for entity linking system's evaluations. As future work we plan to extend this paper's experimental setup and build a ground truth for temporal entity linking as well as try to create an adaptive entity linking system.

\bibliographystyle{splncs04}
\bibliography{main}

\end{document}